\documentclass[conference]{IEEEtran}
\IEEEoverridecommandlockouts
\usepackage{cite}
\usepackage{amsmath,amssymb,amsfonts}
\usepackage{algorithmic}
\usepackage{graphicx}
\usepackage{textcomp}
\usepackage{xcolor}

\usepackage{caption}
\usepackage{float}
\usepackage{booktabs}
\usepackage{multirow}
\def\BibTeX{{\rm B\kern-.05em{\sc i\kern-.025em b}\kern-.08em
    T\kern-.1667em\lower.7ex\hbox{E}\kern-.125emX}}
\begin{document}

\title{360-GeoGS: Geometrically Consistent Feed-Forward 3D Gaussian Splatting Reconstruction for 360 Images}

\author{
\IEEEauthorblockN{
Jiaqi Yao\IEEEauthorrefmark{1},
Zhongmiao Yan\IEEEauthorrefmark{1},
Jingyi Xu\IEEEauthorrefmark{1},
Songpengcheng Xia\IEEEauthorrefmark{1},
Yan Xiang\IEEEauthorrefmark{1},
Ling Pei\IEEEauthorrefmark{1}\IEEEauthorrefmark{2}
\thanks{Corresponding author: ling.pei@sjtu.edu.cn.}
\thanks{This work was supported in part by the Shanghai Collaborative Innovation Fund (XTCX-CY-2025-002), and in part by the Science and Technology Commission of Shanghai Municipality under Grant No. 24DZ3101300 and No. 24TS1402600.}
}
\IEEEauthorblockA{
\IEEEauthorrefmark{1}
Shanghai Key Laboratory of Navigation and Location Based Services,
Shanghai Jiao Tong University
}
\IEEEauthorblockA{
\IEEEauthorrefmark{2}
State Key Laboratory of Submarine Geoscience,
School of Automation and Intelligent Sensing,\\
Shanghai Jiao Tong University
}
}

\maketitle
\begin{abstract}
3D scene reconstruction is fundamental for spatial intelligence applications such as AR, robotics, and digital twins.
Traditional multi-view stereo struggles with sparse viewpoints or low-texture regions, while neural rendering approaches, though capable of producing high-quality results, require per-scene optimization and lack real-time efficiency.
Explicit 3D Gaussian Splatting (3DGS) enables efficient rendering, but most feed-forward variants focus on visual quality rather than geometric consistency, limiting accurate surface reconstruction and overall reliability in spatial perception tasks.
This paper presents a novel feed-forward 3DGS framework for 360 images, capable of generating geometrically consistent Gaussian primitives while maintaining high rendering quality.
A Depth-Normal geometric regularization is introduced to couple rendered depth gradients with normal information, supervising Gaussian rotation, scale, and position to improve point cloud and surface accuracy.
Experimental results show that the proposed method maintains high rendering quality while significantly improving geometric consistency, providing an effective solution for 3D reconstruction in spatial perception tasks.
\end{abstract}

\begin{IEEEkeywords}
3D Reconstruction, 3D Gaussian Splatting, 360 Image.
\end{IEEEkeywords}
\section{Introduction}\label{Introduction}

3D scene reconstruction aims to recover scene geometry and appearance from multi-view observations and is essential for applications such as autonomous driving, AR/VR, robotic perception, and digital twins. In indoor navigation, accurate and efficient 3D modeling is crucial for spatial perception and localization. Prior research has explored robust sensing and efficient inference in complex environments~\cite{Knowledge-based,Nonconvex,Cost-Effective}, highlighting the importance of balancing accuracy, robustness, and computational efficiency.

Multi-View Stereo (MVS) achieves high-precision reconstruction via multi-view matching and depth estimation, but performance degrades under low-texture conditions. Neural Radiance Fields (NeRF)~\cite{nerf} improve view synthesis but require dense inputs and per-scene optimization. For efficient inference, explicit 3D Gaussian Splatting (3DGS)~\cite{3DGS} represents scenes with Gaussian ellipsoids, enabling fast rendering and gradient-based optimization. Following this approach, feed-forward variants~\cite{pixelsplat,mvsplat} improve inference efficiency and generalization through end-to-end prediction, while still often struggling to maintain geometric consistency. In contrast, optimization-based methods, such as VCR-GauS~\cite{vcr} and NeuSG~\cite{neusg}, incorporate geometric priors and normal constraints to refine scene structure, achieving higher accuracy at the cost of efficiency and generalization.

With the increasing adoption of panoramic cameras, 360 images have become an effective source for sparse-view reconstruction~\cite{wu2023360}, capturing the entire scene in a single shot. Feed-forward panoramic methods focus on rendering quality, but projection distortions and unstable depth estimation often cause structural drift, limiting faithful geometric recovery.

To address these challenges, we propose 360-GeoGS, a feed-forward 3DGS framework for 360 image inputs that incorporates Depth-Normal (D-Normal) regularization. The framework predicts multi-view depth from 360 images and fuses various features, which are then processed by a U-Net to regress pixel-aligned Gaussian primitives. D-Normal regularization is applied in the rendering space to jointly supervise Gaussian position, scale, and orientation. Experiments on multiple panoramic benchmarks demonstrate that our method substantially improves geometric consistency and surface continuity while maintaining high rendering quality, outperforming existing feed-forward panoramic 3DGS methods. Our main contributions are as follows:
\begin{figure*}[ht!]
\captionsetup{justification=raggedright,singlelinecheck=false}
\begin{center}
	\includegraphics[width=1.0\textwidth]{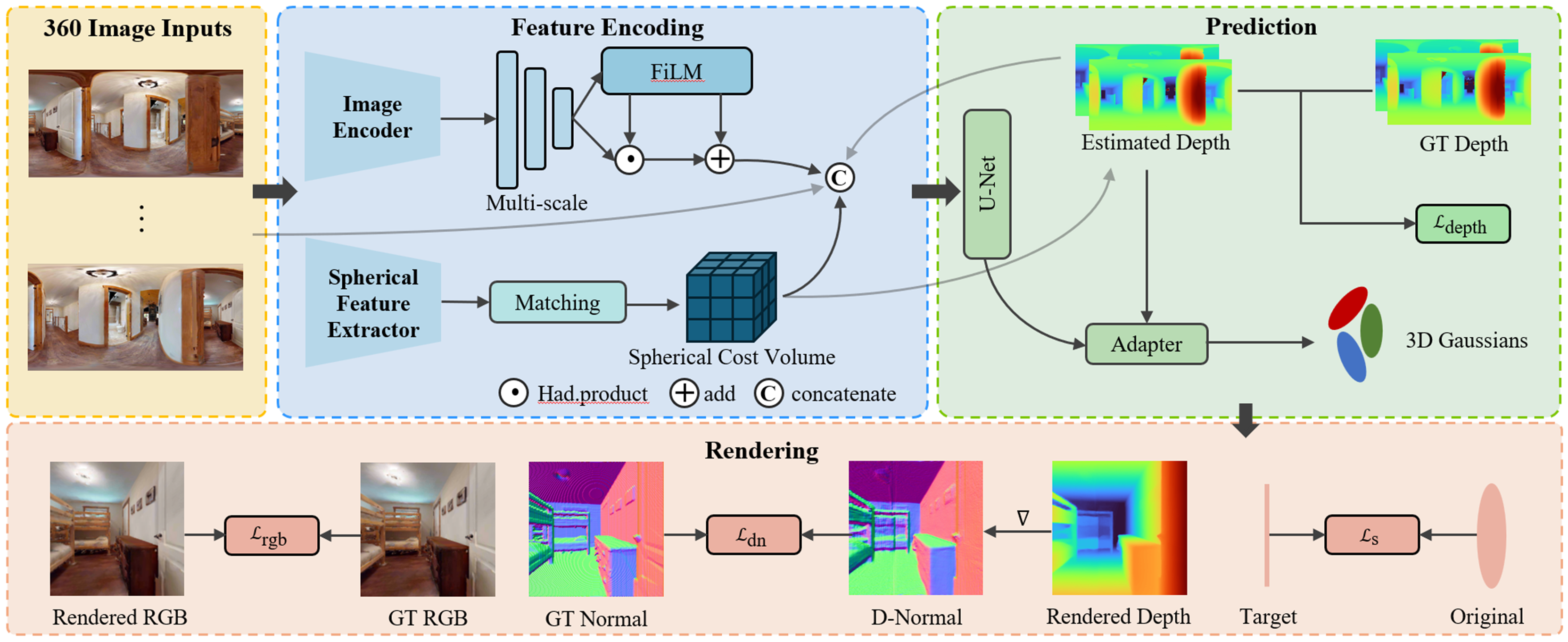}
	\caption{Our pipeline extracts matching features from 360 images using a SphereCNN to construct a spherical cost volume and regress initial depth. Multi-scale features are also extracted by an image encoder and modulated via a FiLM module. The spherical cost volume, modulated multi-scale features, initial RGB images, and depth estimates are then fused to form a unified multi-modal representation, which is decoded by a U-Net and further refined by an adapter to produce per-pixel 3D Gaussian parameters. The network is trained with four losses: $\mathcal{L}_{\mathrm{rgb}}$, $\mathcal{L}_\mathrm{s}$, $\mathcal{L}_{\mathrm{dn}}$, and $\mathcal{L}_{\mathrm{depth}}$ (the definitions of $\mathcal{L}_\mathrm{s}$ and $\mathcal{L}_{\mathrm{dn}}$ are provided in Section~\ref{subsec:D-Normal}).}
\label{fig:pipeline}
\end{center}
\end{figure*}
\begin{enumerate}
\item We propose a feed-forward 3DGS network for 360 image inputs, which employs a SphereCNN backbone to extract spherical features and build a spherical cost volume for depth estimation. Based on the estimated depth, the network performs feed-forward inference to rapidly predict 3D Gaussian parameters, achieving efficient and accurate 3D scene reconstruction.
\item We introduce D-Normal regularization, which jointly optimizes surface normals and Gaussian positions to ensure that neighboring Gaussian primitives form coherent local surfaces with consistent orientation and spatial alignment, thereby enhancing the geometric consistency and accuracy of the predicted 3DGS points.
\item Extensive experiments across multiple benchmarks demonstrate that our approach delivers superior geometric performance while preserving high rendering quality.
\end{enumerate}

\section{Related Work}\label{sec:related work}

\subsection{Sparse View Scene Reconstruction and Synthesis}\label{sec:2-1}

Recent advances in 3D reconstruction and novel view synthesis have been largely driven by NeRF~\cite{nerf} and 3DGS~\cite{3DGS}. Although they were initially designed for dense-view settings, increasing attention has been paid to achieving high-quality reconstruction and synthesis under sparse-view conditions. Existing methods can be divided into per-scene optimization methods~\cite{vcr,neusg,deng2023nerf,ye2024thermal} and cross-scene feed-forward inference methods. The former enhance geometric and appearance stability by designing effective regularization constraints, but the computational cost is high due to the optimization process. In contrast, the latter learn strong priors from large-scale datasets, enabling fast reconstruction through a single forward pass, thus significantly improving inference efficiency.

\subsection{Feed-Forward 3DGS}\label{sec:2-2}

3DGS leverages rasterization-based splatting to efficiently synthesize novel views, representing a scene with learnable Gaussian primitives.
To further accelerate reconstruction and handle sparse-view settings, feed-forward 3DGS variants have been proposed. PixelSplat~\cite{pixelsplat} introduced a feed-forward framework for scene-level Gaussian prediction.
MVSplat~\cite{mvsplat} enhanced geometric accuracy through cost volumes, and DepthSplat~\cite{depthsplat} enhances multi-view consistency with depth estimation.
Despite these advances, feed-forward methods often lack geometric consistency, particularly at indoor scene boundaries with discontinuous depth.
Moreover, most existing methods are designed for perspective images, and their performance degrades significantly on panoramic inputs due to the wide field of view and projection distortions.

\subsection{Panoramic View Scene Reconstruction and Synthesis}\label{sec:2-3}
Reconstruction and novel view synthesis from 360 images encounter challenges caused by geometric distortions in equirectangular projection and unstable depth estimation at high resolutions. Most methods assume dense panoramic inputs~\cite{360gs}, while sparse views make depth and geometry estimation harder.
360Recon~\cite{360recon} predicts panoramic depth using an improved MVS approach, achieving accurate mesh geometry but limited rendering quality. PanoGRF~\cite{panogrf} aggregates geometric and appearance features for high-quality synthesis; however, its large fusion network restricts inference and rendering speed.

Feed-forward 3DGS methods have been extended to 360 images, improving efficiency in panoramic view scene reconstruction and synthesis. Splatter-360~\cite{splatter360}, based on MVSplat, adds depth constraints to enhance geometry but exhibits inconsistencies near scene boundaries. PanSplat~\cite{pansplat} enables high-resolution, real-time synthesis; nevertheless, its geometric constraints are insufficient to fully preserve 3D structure.
\begin{figure*}[tp!]
\begin{center}
		\includegraphics[width=\textwidth]{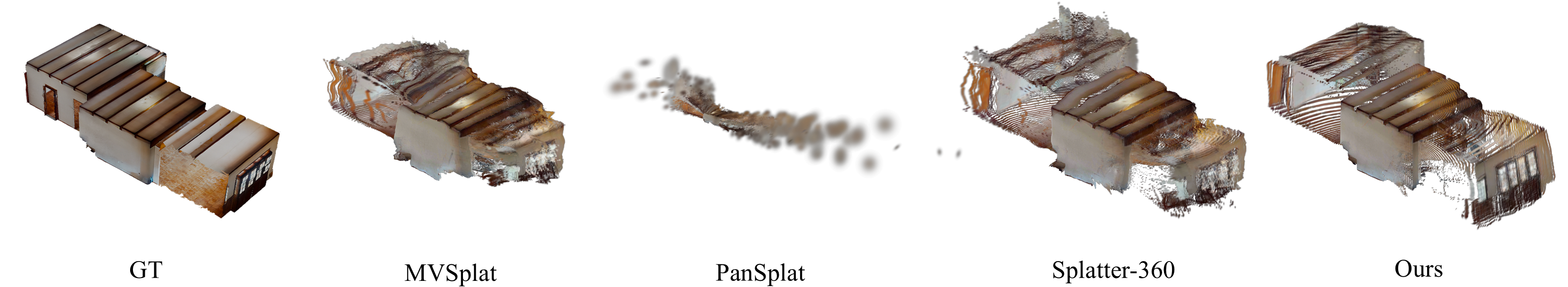}
	\caption{Predicted 3D Gaussian spatial distributions of the same scene reconstructed by different methods.}
\label{fig:GS}
\end{center}
\end{figure*}

\section{Method}\label{sec:method}

Our goal is to directly predict 3DGS parameters from 360 image inputs, enabling geometrically consistent scene reconstruction in a feed-forward manner. 
Section~\ref{subsec:framework} introduces the overall architecture, Section~\ref{subsec:pipeline} details the feed-forward prediction pipeline, and Section~\ref{subsec:D-Normal} presents the proposed D-Normal constraint that enforces geometric consistency.

\subsection{Framework Overview}\label{subsec:framework}
Our network employs a feed-forward design mapping 360 images to 3D Gaussian primitives. 
As illustrated in Fig.~\ref{fig:pipeline}, reconstruction starts with extracting multi-view matching features to build a spherical cost volume, which is then used to estimate an initial dense depth map as a geometric prior. 
Simultaneously, multi-scale features are extracted and modulated through Feature-wise Linear Modulation (FiLM) for cross-scale interaction.
The fused multi-modal features, together with RGB inputs and the depth prior, are processed by a U-Net decoder and an adapter to regress per-pixel Gaussian parameters, including positions, covariance, opacity, and color. 
Predicted Gaussians are jointly supervised by geometric and photometric losses, with geometric supervision emphasizing surface consistency via D-Normal, ensuring accurate local geometry.

\subsection{Pipeline of Feed-forward 3DGS Prediction}\label{subsec:pipeline}

\subsubsection{Feature Encoding}\label{sec:feature}

We adopt the 360Recon framework as our baseline for feature extraction on spherical inputs. 
A SphereCNN backbone is used to obtain matching features from 360 images, which are used to construct a spherical cost volume and estimate an initial dense depth map as a geometric prior. 
Meanwhile, a set of multi-scale feature maps $\{{F_i}\}_{i=0}^{4}$ is extracted, where low-level features retain detailed geometry and high-level features capture global context.
To facilitate interaction across scales, we employ FiLM to adaptively modulate multi-scale features.
Specifically, high-level features are first aggregated to form a global conditioning representation:
\begin{equation}\label{equ:1}
    C_{\mathrm{cond}} = \Phi\left(\{ F_i \}_{i=2}^{4} \right),
\end{equation}
where $\Phi(\cdot)$ denotes the compression and aggregation operation. 
Then, the low-level features are modulated as:
\begin{equation}\label{equ:2}
    \hat{F} = \gamma(C_{\mathrm{cond}}) \cdot F + \beta(C_{\mathrm{cond}}),
\end{equation}
where $\gamma(\cdot)$ and $\beta(\cdot)$ generate per-channel scaling and shifting parameters. 
The fused features are combined with the matching features, dense depth predictions, and RGB inputs to form a multi-modal representation for subsequent 3DGS regression.

\subsubsection{3DGS Parameter Prediction}\label{sec:prediction}

The fused features are decoded by a U-Net decoder to produce initial Gaussian primitive parameters, which are then refined through an adapter module for rendering compatibility.
The adapter normalizes rotations, adjusts scales with depth, and transforms spherical harmonic coefficients to yield per-pixel Gaussian primitives at full resolution (512×1024) aligned with the input panoramas.
The predicted parameters include:

\textbf{Gaussian centers $\mu$.} 
The network predicts per-pixel offsets in image space, which are combined with depth to project points into 3D camera coordinates and further transformed to world coordinates using the camera-to-world matrix.

\textbf{Opacity $\alpha$.} 
Opacity is derived from the matching confidence, computed as a normalized probability distribution from the cost volume.

\textbf{Covariance $\Sigma$.} 
The covariance is defined by a scale factor $s$ and rotation matrix $R(\theta)$:
\begin{equation}\label{equ:5}
	\Sigma = R(\theta)^T \text{diag}(s) R(\theta),
\end{equation}
where $s$ is mapped through a Sigmoid function to preserve proportionality to depth and image resolution, and $R(\theta)$ is parameterized via a normalized quaternion.

\textbf{Spherical harmonics $c$.} 
The spherical harmonic coefficients $c$ are regressed from the fused features to encode view-dependent color representations.

\begin{figure*}[ht!]
\begin{center}
		\includegraphics[width=\textwidth]{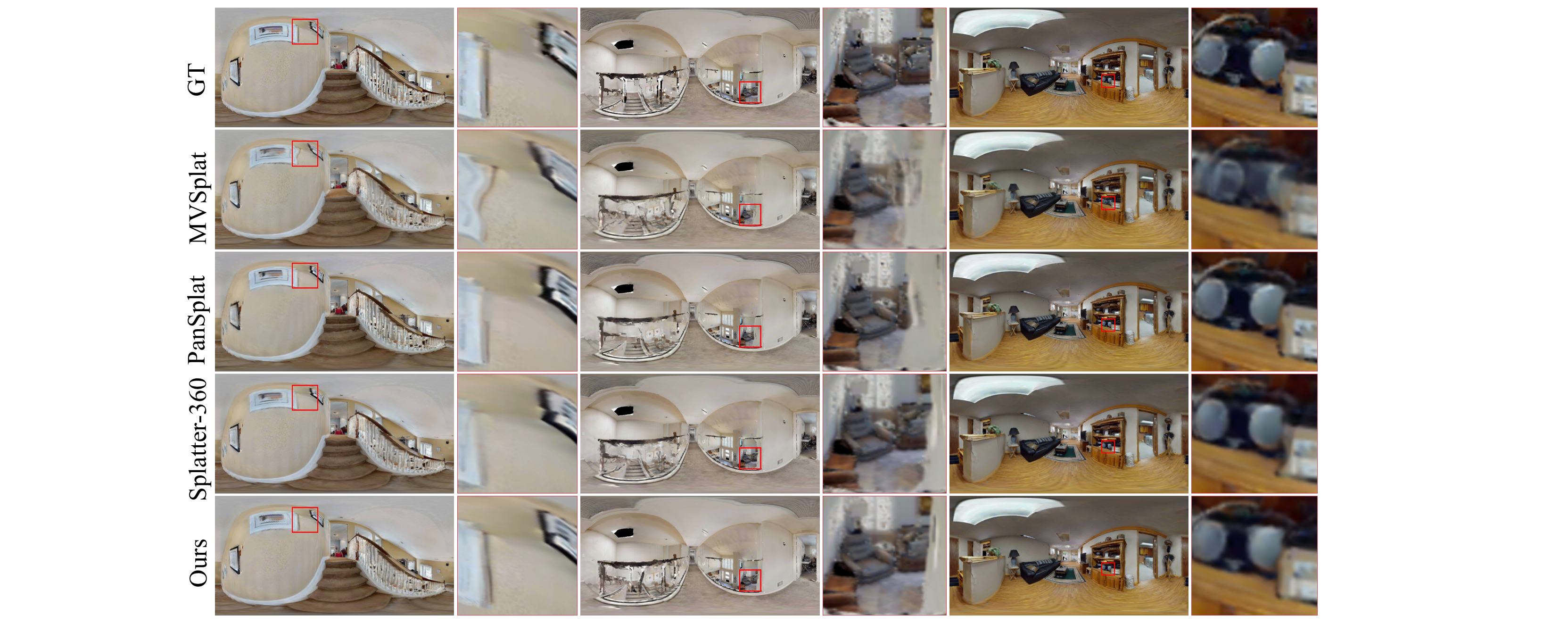}
	\caption{Novel view rendering comparison of our method, Splatter-360, PanSplat, and MVSplat on the HM3D dataset.}
\label{fig:figure_placement}
\end{center}
\end{figure*}

\begin{figure*}[ht!]
\begin{center}
		\includegraphics[width=\textwidth]{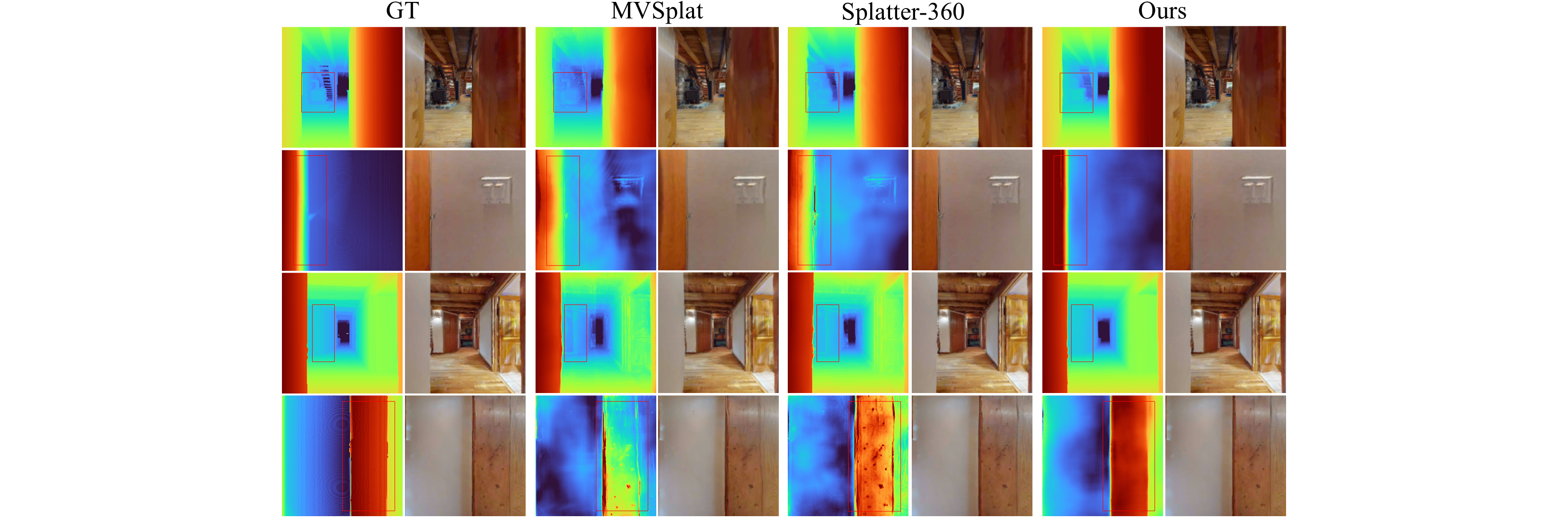}
	\caption{Novel view depth comparison among MVSplat, Splatter-360, and our method on the HM3D dataset.}
\label{fig:depth}
\end{center}
\end{figure*}

\subsection{Geometric Constraint}\label{subsec:D-Normal}
To enhance the geometric accuracy of feed-forward 3DGS predictions and better align Gaussian points with object surfaces, we introduce a geometric constraint.

\renewcommand{\arraystretch}{0.75} 
\begin{table*}[htbp]
  \centering  \captionsetup{justification=raggedright,singlelinecheck=false}
  \scriptsize    
  \caption{Quantitative comparison of depth estimation metrics on the HM3D and Replica datasets. $^\dagger$ indicates the model that was trained by us on the panoramic dataset. Best in each column is bolded.}
  \label{tab:depth}
  \begin{tabular}{lcccccccc}
    \toprule
    \multirow{2}{*}{Method}  & \multicolumn{4}{c}{HM3D} & \multicolumn{4}{c}{Replica} \\
    \cmidrule(lr){2-5} \cmidrule(lr){5-9}
           & Abs Diff$\downarrow$ & Abs Rel$\downarrow$ & RMSE$\downarrow$ & $\delta < 1.25$ $\uparrow$
           & Abs Diff$\downarrow$ & Abs Rel$\downarrow$ & RMSE$\downarrow$ & $\delta < 1.25$ $\uparrow$ \\
    \midrule
    MVSplat$^\dagger$ & 0.140 & 0.094 & 0.258 & 91.150 & 0.186 & 0.111 & 0.282 & 88.216 \\
    Splatter-360  & 0.098 & \textbf{0.068} & 0.193 & 95.417 & 0.103 & \textbf{0.068} & 0.185 & 95.412\\
    Ours & \textbf{0.053} & 0.069 & \textbf{0.141} & \textbf{96.423} & \textbf{0.055} & \textbf{0.068} & \textbf{0.138} & \textbf{96.528} \\
    \bottomrule
  \end{tabular}
\end{table*}

\subsubsection{Normal and Intersection Depth}\label{sec:Normal and Depth}
The spatial positions of feed-forward 3DGS points primarily depend on the estimated depth and are theoretically expected to lie on object surfaces. However, since Gaussians are represented as ellipsoids, their centers often deviate from the true surface, leading to geometric inconsistencies. To address this, we follow NeuSG and compress each ellipsoid along its smallest scale direction into a height-flattened form, allowing the Gaussian to better adhere to the underlying surface.

\renewcommand{\arraystretch}{0.75} 
\begin{table*}[htbp]
  \scriptsize
  \centering
    \caption{Quantitative comparison of novel view synthesis metrics on the HM3D and Replica datasets. Best in each column is bolded.}
  \label{tab:render}
  \setlength{\tabcolsep}{13pt}  
  \begin{tabular}{lcccccc}
    \toprule
    \multirow{2}{*}{Method}  & \multicolumn{3}{c}{HM3D} & \multicolumn{3}{c}{Replica} \\
    \cmidrule(lr){2-4} \cmidrule(lr){5-7}
           & PSNR$\uparrow$ & SSIM$\uparrow$ & LPIPS$\downarrow$
           & PSNR$\uparrow$ & SSIM$\uparrow$ & LPIPS$\downarrow$ \\
    \midrule
    MVSplat$^\dagger$ & 29.537 & 0.892 & 0.138  & 28.682 & 0.915 & 0.117 \\ 
    PanSplat     & 29.733 & \textbf{0.925} & 0.126 & \textbf{31.821} & \textbf{0.960} & 0.067 \\
    Splatter-360  & \textbf{31.669} & \textbf{0.925} & 0.100 & 31.584 & 0.952 & \textbf{0.064} \\
    Ours & 31.043 & 0.920 & \textbf{0.098} & 31.137 & 0.945 & 0.066 \\
    \bottomrule
  \end{tabular}
  \setlength{\tabcolsep}{6pt}
\end{table*}

Specifically, the scale factor $\mathbf{s}=(s_1,s_2,s_3)^T$ defines the ellipsoid’s extent along each principal axis. The normal vector $\mathbf{n}$ is then defined along the direction of the minimal scale component. Minimizing this component effectively flattens the ellipsoid, and a scale regularization loss $\mathcal{L}_\mathrm{s}$ is applied to constrain it towards zero:
\begin{equation}\label{equ:6}
\mathcal{L}_\mathrm{s}=\|\min(s_1,s_2,s_3)\|_1.\end{equation}

In depth computation, conventional methods typically obtain the depth from the center position $\textbf{p}=(p_x,p_y,p_z)$ of each Gaussian in the camera coordinate system. However, this ignores the normal vector $\textbf{n}$ and thus limits the effectiveness of geometric constraints. We therefore adopt a more appropriate approach, computing the intersection depth between the camera ray $\textbf{r}$ and the flattened Gaussian surface, defined as:
\begin{equation}\label{equ:7}
\mathbf{d}(\mathbf{n},\mathbf{p})={r_z(\mathbf{n}\cdot \mathbf{p})}/{(\mathbf{n}\cdot \mathbf{r})},
\end{equation}
Here, $r_z$ denotes the z-component of the ray $\textbf{r}$. The intersection depth depends on both the position $\textbf{p}$ and the normal vector $\textbf{n}$ of the Gaussian, allowing them to be jointly constrained during optimization to improve depth estimation accuracy.

\subsubsection{D-Normal Regularization}\label{sec:D-Normal formula}
\begin{figure}[tbp]
  \centering  \captionsetup{width=\columnwidth,justification=justified,singlelinecheck=false}
  \includegraphics[width=8cm]{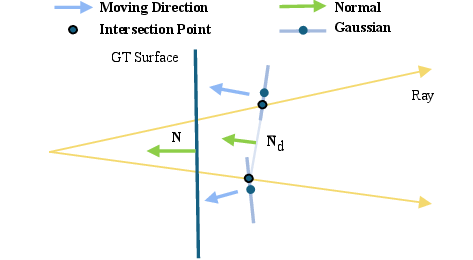}
  \caption{Illustration of the D-Normal regularization. $\bar{\mathbf{N}}_{d}$ is supervised by the ground-truth normal through $\mathcal{L}_{\mathrm{dn}}$(defined in subsection~\ref{sec:D-Normal formula} ), guiding the flattened Gaussians to better fit the true surface.}
  \label{fig:surface}
\end{figure}

Following this approach, we adopt the D-Normal regularization. Specifically, a depth map is generated using the 3DGS renderer, following a procedure analogous to RGB rendering.
\begin{equation}
\label{equ:8}
\hat{D}={\sum_{i\in M} d_i\,\alpha_i\, T_i}/({\sum_{i\in M} \alpha_i\, T_i})
\qquad\quad
T_i=\prod_{j=1}^{i-1}(1-\alpha_j)
\end{equation}
where $d_i$ is the intersection depth from \eqref{equ:7} and $M$ is the number of Gaussians the ray passes through. Subsequently, the rendered normal 
$\bar{\textbf{N}}_{d}(\textbf{n},\textbf{p})$ is obtained by computing finite differences of the depth map along the horizontal and vertical directions and taking their cross product. This normal depends on both the Gaussian normal $\textbf{n}$ and the position 
$\textbf{p}$:
\begin{equation}\label{equ:9}
\bar{\mathbf{N}}_{d}(\mathbf{n},\mathbf{p})=\frac{\nabla_{v}\mathbf{d}(\mathbf{n},\mathbf{p})\times\nabla_{h}\mathbf{d}(\mathbf{n},\mathbf{p})}{|\nabla_{v}\mathbf{d}(\mathbf{n},\mathbf{p})\times\nabla_{h}\mathbf{d}(\mathbf{n},\mathbf{p})|}.
\end{equation}
The D-Normal regularization enforces consistency between the rendered normal $\bar{\mathbf{N}}_{d}$ and the target normal $\mathbf{N}$, enabling joint optimization of Gaussian positions and orientations, as illustrated in Fig.~\ref{fig:surface}. The regularization loss is formulated as:
\begin{equation}\label{equ:10}
\mathcal{L}_{\mathrm{dn}}=\|\bar{\mathbf{N}}_{d}-\mathbf{N}\|_{1}+(1-\bar{\mathbf{N}}_{d}\cdot\mathbf{N}).
\end{equation}
Our overall loss function is defined as:
\begin{equation}
\mathcal{L}_{\mathrm{total}}=\mathcal{L}_{\mathrm{rgb}}+\lambda_{1}\mathcal{L}_{\mathrm{s}}+\lambda_{2}\mathcal{L}_{\mathrm{depth}}+\lambda_{3}\mathcal{L}_{\mathrm{dn}}.
\end{equation}

\section{Experiments}\label{sec:experiments}

\subsection{Implementation Details}\label{subsec:details}Our method is implemented in PyTorch, with intersection distance computations accelerated using custom CUDA kernels. All experiments are conducted on a single NVIDIA A100 GPU with 80 GB VRAM. The RGB loss for training is a linear combination of MSE and LPIPS, weighted 1 and 0.05, respectively, and the hyperparameters $\lambda_1$, $\lambda_2$, and $\lambda_3$ are empirically set to 1, 0.1, and 0.01.

\subsection{Datasets and Metrics}\label{subsec:datasets}We evaluate our model on two panoramic datasets, HM3D~\cite{hm3d} and Replica~\cite{replica}, which contain diverse indoor scenes. For quantitative comparison, we compare our method with several state-of-the-art 360 approaches, including Splatter-360 and PanSplat. Additionally, MVSplat is retrained on the 360 datasets and included as another baseline. We evaluate the performance of all methods on novel view synthesis using standard metrics, including PSNR, SSIM, and LPIPS, and on depth prediction using Abs Diff, Abs Rel, RMSE, and $\delta < 1.25$. Moreover, geometric reconstruction is assessed on HM3D using Accuracy, Completeness, and Chamfer Distance.

\renewcommand{\arraystretch}{0.70}
\begin{table*}[htbp]
  \centering  
  \scriptsize
    \caption{Quantitative results of the ablation study. ``w/o'' indicates ``without''. Best in each column is bolded. }
  \label{tab:ablation_full}
  \setlength{\tabcolsep}{3pt}
  \begin{tabular}{lccc|ccccccc}
    \toprule
    Method & PSNR$\uparrow$ & SSIM$\uparrow$ & LPIPS$\downarrow$ & 
    Abs Diff$\downarrow$ & Abs Rel$\downarrow$ & RMSE$\downarrow$ & $\delta<1.25\uparrow$ &
    Acc(m)$\downarrow$ & Comp(m)$\downarrow$ & Chamfer(m)$\downarrow$ \\
    \midrule
    w/o D-N & 29.078 & 0.887 & 0.161 & 0.086 & \textbf{0.068} & 0.177 & 94.878 & 0.054 & 0.715 & 0.769 \\
    w/o D-N+Scales & 28.189 & 0.868 & 0.158 & 0.111 & 0.102 & 0.205 & 93.715 & 0.059 & 0.731 & 0.790 \\
    w/o D-N+Scales+Fusion & 27.713 & 0.841 & 0.176 & 0.120 & 0.108 & 0.228 & 93.430 & 0.061 & 0.742 & 0.802 \\
    Full & \textbf{31.043} & \textbf{0.920} & \textbf{0.098} & \textbf{0.053} & 0.069 & \textbf{0.141} & \textbf{96.423} & \textbf{0.049} & \textbf{0.691} & \textbf{0.740} \\
    \bottomrule
  \end{tabular}
\end{table*}

\renewcommand{\arraystretch}{0.70}
\setlength{\tabcolsep}{10pt}
\begin{table}[!t]
  \centering
  \scriptsize  \captionsetup{justification=raggedright,singlelinecheck=false}
    \caption{Quantitative comparison of 3D reconstruction metrics on the HM3D dataset. Best in each column is bolded.}
  \label{tab:reconstruction}  
  \begin{tabular}{lccc}
    \toprule
    Method & Acc(m)$\downarrow$ & Comp(m)$\downarrow$ & Chamfer(m)$\downarrow$ \\
    \midrule
    MVSplat$^\dagger$  & 0.076 & 0.862 & 0.938 \\
    Splatter-360  & 0.062 & 0.719 & 0.780 \\
     Ours & \textbf{0.049} & \textbf{0.691} & \textbf{0.740} \\
    \bottomrule
  \end{tabular}
\end{table}

\subsection{Qualitative Results}\label{subsec:qualitative}Qualitative comparisons are presented in Fig.~\ref{fig:GS}, Fig.~\ref{fig:figure_placement}, and Fig.~\ref{fig:depth}. As Fig.~\ref{fig:figure_placement} illustrates, state-of-the-art methods achieve visually similar results for novel view synthesis, with our method and Splatter-360 producing slightly better appearance in the right-side sample. In Fig.~\ref{fig:depth}, MVSplat exhibits notable depth errors, such as the left edge of the door in Sample 2, while Splatter-360 improves overall reconstruction but still shows limited surface depth consistency in Samples 2 and 4. In contrast, our method generates depth predictions with stronger geometric consistency and higher accuracy. The predicted 3DGS point clouds in Fig.~\ref{fig:GS} further highlight this improvement, demonstrating that our approach produces 3DGS points with clearly enhanced surface.

\subsection{Quantitative Results}\label{subsec:quantitative}

Tables~\ref{tab:depth},~\ref{tab:render}, and~\ref{tab:reconstruction} summarize the quantitative performance of our method compared with MVSplat, Splatter-360, and PanSplat on the HM3D and Replica datasets. As shown in Tables~\ref{tab:depth} and~\ref{tab:reconstruction}, our approach outperforms Splatter-360 in geometric reconstruction and depth estimation metrics, indicating that the predicted 3DGS points exhibit stronger geometric consistency and better alignment with object surfaces. Table~\ref{tab:render} reports novel view synthesis metrics, where our method achieves rendering quality comparable to the current state-of-the-art, with minor differences across multiple metrics. Overall, these results demonstrate that our feed-forward 3DGS framework achieves high-fidelity geometric reconstruction while maintaining competitive rendering performance.

\subsection{Ablation Results}\label{subsec:ablation}
We conduct an ablation study to evaluate the contributions of D-Normal (D-N), scale flattening (Scales), and multi-scale feature fusion (Fusion). As shown in Table~\ref{tab:ablation_full}, the full model consistently outperforms the ablated variants across rendering, depth, and geometric reconstruction metrics. Removing D-N or scale flattening degrades depth accuracy and geometric consistency, while omitting multi-scale fusion reduces rendering quality. These results indicate that each component contributes complementarily, with the integrated model producing the most accurate and geometrically consistent 3DGS predictions.

\section{Conclusion}\label{sec:conclusion}
In this paper, we propose a feed-forward 3DGS framework for 360 image inputs, integrating multi-view matching features, multi-scale feature encoding with FiLM, depth priors from a spherical cost volume, and D-Normal regularization. The encoded features are decoded by a U-Net and refined via an adapter to produce per-pixel Gaussian primitive parameters. Our method enables accurate scene reconstruction and high-fidelity novel view synthesis under sparse-view conditions. Experiments demonstrate competitive rendering quality, precise depth estimation, and enhanced geometric consistency compared to state-of-the-art methods, while ablation studies confirm the effectiveness of each component.

\textbf{Limitations and future work.} Our approach currently targets indoor scenes and relies on accurate camera poses. Future work will explore pose-free reconstruction from 360 images and investigate whether occluded regions can be recovered using generative models, further enhancing the completeness and realism of panoramic scene reconstruction.
\bibliographystyle{IEEEtran}  
\bibliography{references} 

\begin{thebibliography}{10}
\providecommand{\url}[1]{#1}
\csname url@samestyle\endcsname
\providecommand{\newblock}{\relax}
\providecommand{\bibinfo}[2]{#2}
\providecommand{\BIBentrySTDinterwordspacing}{\spaceskip=0pt\relax}
\providecommand{\BIBentryALTinterwordstretchfactor}{4}
\providecommand{\BIBentryALTinterwordspacing}{\spaceskip=\fontdimen2\font plus
\BIBentryALTinterwordstretchfactor\fontdimen3\font minus \fontdimen4\font\relax}
\providecommand{\BIBforeignlanguage}[2]{{%
\expandafter\ifx\csname l@#1\endcsname\relax
\typeout{** WARNING: IEEEtran.bst: No hyphenation pattern has been}%
\typeout{** loaded for the language `#1'. Using the pattern for}%
\typeout{** the default language instead.}%
\else
\language=\csname l@#1\endcsname
\fi
#2}}
\providecommand{\BIBdecl}{\relax}
\BIBdecl

\bibitem{Knowledge-based}
Y.~Chen, R.~Chen, L.~Pei, T.~Kröger, H.~Kuusniemi, J.~Liu, and W.~Chen, ``Knowledge-based error detection and correction method of a multi-sensor multi-network positioning platform for pedestrian indoor navigation,'' in \emph{IEEE/ION Position, Location and Navigation Symposium}, 2010, pp. 873--879.

\bibitem{Nonconvex}
F.~Wen, L.~Adhikari, L.~Pei, R.~F. Marcia, P.~Liu, and R.~C. Qiu, ``Nonconvex regularization-based sparse recovery and demixing with application to color image inpainting,'' \emph{IEEE Access}, vol.~5, pp. 11\,513--11\,527, 2017.

\bibitem{Cost-Effective}
Y.~Li, K.~Yan, Z.~He, Y.~Li, Z.~Gao, L.~Pei, R.~Chen, and N.~El-Sheimy, ``Cost-effective localization using rss from single wireless access point,'' \emph{IEEE Transactions on Instrumentation and Measurement}, vol.~69, no.~5, pp. 1860--1870, 2020.

\bibitem{nerf}
B.~Mildenhall, P.~P. Srinivasan, M.~Tancik, J.~T. Barron, R.~Ramamoorthi, and R.~Ng, ``Nerf: Representing scenes as neural radiance fields for view synthesis,'' \emph{Communications of the ACM}, vol.~65, no.~1, pp. 99--106, 2021.

\bibitem{3DGS}
B.~Kerbl, G.~Kopanas, T.~Leimkuehler, and G.~Drettakis, ``3d gaussian splatting for real-time radiance field rendering,'' \emph{ACM Trans. Graph.}, vol.~42, no.~4, July 2023.

\bibitem{pixelsplat}
D.~Charatan, S.~L. Li, A.~Tagliasacchi, and V.~Sitzmann, ``pixelsplat: 3d gaussian splats from image pairs for scalable generalizable 3d reconstruction,'' in \emph{Proceedings of the IEEE/CVF conference on computer vision and pattern recognition}, 2024, pp. 19\,457--19\,467.

\bibitem{mvsplat}
Y.~Chen, H.~Xu, C.~Zheng, B.~Zhuang, M.~Pollefeys, A.~Geiger, T.-J. Cham, and J.~Cai, ``Mvsplat: Efficient 3d gaussian splatting from sparse multi-view images,'' in \emph{European Conference on Computer Vision}.\hskip 1em plus 0.5em minus 0.4em\relax Springer, 2024, pp. 370--386.

\bibitem{vcr}
H.~Chen, F.~Wei, C.~Li, T.~Huang, Y.~Wang, and G.~H. Lee, ``Vcr-gaus: view consistent depth-normal regularizer for gaussian surface reconstruction,'' in \emph{Proceedings of the 38th International Conference on Neural Information Processing Systems}, 2024.

\bibitem{neusg}
H.~Chen, C.~Li, and G.~H. Lee, ``Neusg: Neural implicit surface reconstruction with 3d gaussian splatting guidance,'' \emph{CoRR}, vol. abs/2312.00846, 2023.

\bibitem{wu2023360}
Q.~Wu, X.~Xu, X.~Chen, L.~Pei, C.~Long, J.~Deng, G.~Liu, S.~Yang, S.~Wen, and W.~Yu, ``360-vio: A robust visual--inertial odometry using a 360 camera,'' \emph{IEEE Transactions on Industrial Electronics}, vol.~71, no.~9, pp. 11\,136--11\,145, 2023.

\bibitem{deng2023nerf}
J.~Deng, Q.~Wu, X.~Chen, S.~Xia, Z.~Sun, G.~Liu, W.~Yu, and L.~Pei, ``Nerf-loam: Neural implicit representation for large-scale incremental lidar odometry and mapping,'' in \emph{Proceedings of the IEEE/CVF International Conference on Computer Vision}, 2023, pp. 8218--8227.

\bibitem{ye2024thermal}
T.~Ye, Q.~Wu, J.~Deng, G.~Liu, L.~Liu, S.~Xia, L.~Pang, W.~Yu, and L.~Pei, ``Thermal-nerf: Neural radiance fields from an infrared camera,'' in \emph{2024 IEEE/RSJ International Conference on Intelligent Robots and Systems (IROS)}.\hskip 1em plus 0.5em minus 0.4em\relax IEEE, 2024, pp. 1046--1053.

\bibitem{depthsplat}
H.~Xu, S.~Peng, F.~Wang, H.~Blum, D.~Barath, A.~Geiger, and M.~Pollefeys, ``Depthsplat: Connecting gaussian splatting and depth,'' in \emph{Proceedings of the Computer Vision and Pattern Recognition Conference}, 2025, pp. 16\,453--16\,463.

\bibitem{360gs}
J.~Bai, L.~Huang, J.~Guo, W.~Gong, Y.~Li, and Y.~Guo, ``360-gs: Layout-guided panoramic gaussian splatting for indoor roaming,'' \emph{CoRR}, vol. abs/2402.00763, 2024.

\bibitem{360recon}
Z.~Yan, Q.~Wu, S.~Xia, J.~Deng, X.~Mu, R.~Jin, and L.~Pei, ``360recon: An accurate reconstruction method based on depth fusion from 360 images,'' \emph{CoRR}, vol. abs/2411.19102, 2024.

\bibitem{panogrf}
Z.~Chen, Y.-P. Cao, Y.-C. Guo, C.~Wang, Y.~Shan, and S.-H. Zhang, ``Panogrf: Generalizable spherical radiance fields for wide-baseline panoramas,'' \emph{Advances in Neural Information Processing Systems}, vol.~36, pp. 6961--6985, 2023.

\bibitem{splatter360}
Z.~Chen, C.~Wu, Z.~Shen, C.~Zhao, W.~Ye, H.~Feng, E.~Ding, and S.-H. Zhang, ``Splatter-360: Generalizable 360 gaussian splatting for wide-baseline panoramic images,'' in \emph{Proceedings of the IEEE/CVF Conference on Computer Vision and Pattern Recognition (CVPR)}, June 2025, pp. 21\,590--21\,599.

\bibitem{pansplat}
C.~Zhang, H.~Xu, Q.~Wu, C.~C. Gambardella, D.~Phung, and J.~Cai, ``Pansplat: 4k panorama synthesis with feed-forward gaussian splatting,'' in \emph{Proceedings of the IEEE/CVF Conference on Computer Vision and Pattern Recognition}, 2025.

\bibitem{hm3d}
S.~K. Ramakrishnan, A.~Gokaslan, E.~Wijmans, O.~Maksymets, A.~Clegg, J.~Turner, E.~Undersander, W.~Galuba, A.~Westbury, A.~X. Chang \emph{et~al.}, ``Habitat-matterport 3d dataset (hm3d): 1000 large-scale 3d environments for embodied ai,'' \emph{CoRR}, vol. abs/2109.08238, 2021.

\bibitem{replica}
J.~Straub, T.~Whelan, L.~Ma, Y.~Chen, E.~Wijmans, S.~Green, J.~J. Engel, R.~Mur-Artal, C.~Ren, S.~Verma \emph{et~al.}, ``The replica dataset: A digital replica of indoor spaces,'' \emph{CoRR}, vol. abs/1906.05797, 2019.

\end{thebibliography}
\end{document}